\documentclass[10pt]{article}

\usepackage[margin=1in]{geometry}
\usepackage{amsmath,amssymb,amsthm}
\usepackage{booktabs}
\usepackage{microtype}
\usepackage{xcolor}
\usepackage{graphicx}
\usepackage{float}
\usepackage{enumitem}
\usepackage[hidelinks]{hyperref}
\usepackage{caption}

\usepackage[T1]{fontenc}
\usepackage{lmodern}
\usepackage{textcomp}

\theoremstyle{definition}
\newtheorem{definition}{Definition}
\newtheorem{proposition}{Proposition}

\newcommand{\Step}{\mathcal{M}}
\newcommand{\View}{\operatorname{View}}
\newcommand{\Edit}{\operatorname{Edit}}
\newcommand{\Undo}{\operatorname{Undo}}

\title{\textbf{Transfiver: Human--AI Co-Inference through a Shared Editable State}}

\author{
  \textbf{Minji Park},
  \textbf{Seunghyun Yoon},
  \textbf{Hyuk Lim} \\[0.5em]
  Korea Institute of Energy Technology (KENTECH) \\[0.3em]
  \small{\texttt{\{minjipark, syoon, hlim\}@kentech.ac.kr}}
}

\begin{document}

\maketitle

\begin{abstract}
Long-term human--AI interaction is difficult because the information that guides inference is updated implicitly by the model and is not directly inspectable or controllable by the user. We introduce the TRANSparent Framework for Interactive, Verifiable, Editable Representation (Transfiver), an architecture for human--AI co-inference through a shared editable state. Its central idea is that interaction-specific information is maintained in a single persistent state \(S_t\) that both the model and the human update.

Transfiver distinguishes two modes of state evolution. In an implicit stream update, the model interprets ongoing interaction and decides whether new information revises an existing state item or creates a new one. In an explicit directed edit, a human inspects and modifies an addressed item. Both act on the same underlying state, so a human correction changes the state that subsequent computation reads, rather than adding another instruction or separate record.

The architecture separates shared parameters \(\theta\), learned before ordinary use, from the persistent state \(S_t\), which evolves during deployment without parameter retraining. Extending Transfiver to rich natural-language, relational, and large-scale shared states remains open.
\end{abstract}

\section{Introduction}
\label{sec:intro}

When an assistant is used over months, some of what it was told stops being true.
A meeting moves, a password is retired, a rule that held for one project is not
meant for another. Somebody has to decide that an old entry no longer counts, and
the question this paper asks is who that somebody is.

The prevailing answer is the model. Recent work trains agents to keep the current
value of a changed fact and to stop answering from stale ones, and reports that
this is both a real failure and a trainable one~\cite{supersede2026}. The person is the one who knows that the meeting moved,
and the correction they make should be an operation on the state the assistant
actually reads, not another sentence added to a context window that a later turn
may reassemble without it.

This requires the memory a user sees and edits to be the same state $S_t$ the
model computes from. Transfiver makes that identity the architectural
requirement: interaction-specific information lives in one persistent state that
both parties update, implicitly when the model folds an ongoing interaction into
it, explicitly when a person retires, restores or rewrites an addressed entry.
Omitting an entry when assembling one turn's context looks like retiring it, and
on that turn the model sees exactly the same lines. The difference appears on the
next turn, when the omitted entry returns and the retired one does not.

\begin{figure}[t]
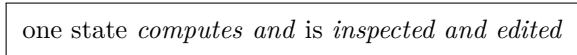

\centering
\setlength{\fboxsep}{6pt}
\begin{tabular}{c}
\textbf{A permitted failure: two separate states}\\[5pt]
\fbox{state that \emph{computes the answer}}\quad$\longleftrightarrow$\quad\fbox{state you can \emph{see and edit}}\\[3pt]
{\small they can diverge, so a correction need not take effect}\\[12pt]
\textbf{Transfiver: one state}\\[5pt]
\fbox{one state \emph{computes} \emph{and} is \emph{inspected and edited}}\\
\end{tabular}
\caption{Many current designs permit the state a person can inspect and edit to remain separate from the state that produces the next output, so a correction need not take effect. Transfiver binds them into one state: editing what you see edits the computation, and restoring it restores the computation.}
\label{fig:identity}
\end{figure}

When humans interact with a language model over a long time, it accumulates corrections, obsolete beliefs, user-specific meanings, unresolved conflicts, evidence of use, and decisions that may later be revived. Transformer activations encode the current sequence~\cite{vaswani2017attention}; recurrent and state-space models compress the past into latent state~\cite{gu2024mamba}; retrieval systems preserve explicit records outside the generator~\cite{lewis2020rag,borgeaud2022retro}; and agent frameworks move records between memory tiers~\cite{packer2023memgpt}. These approaches do not guarantee that the state a user sees and edits is the only state the model keeps and uses for future answers.

A user may commit an edit, but another hidden computational path may still carry the old information. Transfiver requires:

\begin{quote}
Memory, reasoning, and output are not divided. The state underlying what a person sees and corrects is the state from which the system computes, updates, and answers.
\end{quote}

Transfiver is intended to operate at the same architectural level as a Transformer or a state-space model. It may also be embedded within a larger system, provided that Transfiver remains the sole persistent interaction-specific state through which prior interaction can affect future computation. Trained on language, Transfiver may become a language model. Applied to an information environment, the same architecture may become the interface through which that environment is read and changed.

\paragraph{Contributions.}

\begin{enumerate}[leftmargin=*]
  \item We introduce Transfiver's information model and its human-facing renderings, and specify the distinctions that the state must preserve.
  \item We turn ``memory, reasoning, and output in one living state'' into a history-sufficiency and causal-intervention contract that distinguishes Transfiver from a memory attachment, a post-hoc explanation, or a display the computation can ignore.
  \item We provide a small working implementation that tests some of Transfiver's core principles. It is not intended to represent the full Transfiver architecture.
  \item We show, on an earlier persistent-state implementation, in a single held-out evaluation with controls fixed in advance, that one frozen persistent state retains enough temporal structure to support four disagreeing read rules, and that replacing the named rule or reference redirects the readout accordingly.
\end{enumerate}

\section{Related Work}

We group prior work by the question it answers about a persistent state: who
writes it, who may read it, and whether it is the only path from past interaction
to future computation.

\paragraph{Self-updating computational memory.}
Neural Turing Machines and memory-augmented networks learn to address external
memory~\cite{graves2014ntm,santoro2016mann}.  Fast weights, test-time training,
and Titans add state that changes on a faster time scale than ordinary
parameters~\cite{ba2016fastweights,sun2024ttt,behrouz2025titans}.
Transfiver differs in requiring its persistent state to be directly inspectable and editable, while also serving as the state used for computation.

Transformers retain token-level context through attention
\cite{vaswani2017attention}; selective state-space models learn compact recurrent
dynamics with strong efficiency and sequence performance~\cite{gu2024mamba}. Transfiver's additional requirement is that persistent, addressable, human-viewed state be the causal
computational substrate across interactions.

\paragraph{State-first agent memory.}
Recent work measures the same failure we target from the model's side. Supersede
shows that when an agent maintains a bounded self-managed memory instead of full
context, accuracy on questions about changed facts drops from $.92$ to $.77$, and
that the gap widens with conversation length rather than with compression ratio;
it releases a reinforcement-learning environment that rewards answering from the
current value and penalizes stale ones, and reports that fine-tuning nearly
doubles held-out accuracy~\cite{supersede2026}. That line of work makes the model
a better custodian of its own memory. Transfiver takes the complementary position:
the model may judge which stored item an incoming event concerns, but the
authority to retire, restore and correct that item belongs to the user, and the
item the user acts on is the item subsequent computation reads. Supersede does not
expose the memory for human inspection or editing, and does not test intervention,
rollback, or save--restore.

$k$NN language models, RAG, and RETRO combine a generator with explicit external
records~\cite{khandelwal2020knnlm,lewis2020rag,borgeaud2022retro}; MemGPT manages
memory tiers around an LLM~\cite{packer2023memgpt}; and Larimar adds an editable episodic memory that supports one-shot knowledge updates without retraining the language model for each edit~\cite{das2024larimar}. Memory Layers at Scale is closer in spirit, making a large
addressable key--value store part of the forward computation rather than a
retrieval sidecar~\cite{berges2024memorylayers}. Transfiver treats the
separation between the readable store and the computing model as the object of
study, and requires the readable state to be the sole persistent interaction-specific state on which future computation depends.

ROME and MEMIT intervene on factual associations in Transformer parameters
\cite{meng2022rome,meng2023memit}.  Transfiver instead separates slowly learned shared
dynamics $\theta$ from an online, user- or task-specific $S$.  This avoids
gradient editing during ordinary use, while imposing the stronger burden that
state edits be causally localized and serializable.

\paragraph{Shared mental models in human--agent teams.}
Teaming research has long argued that effective human--agent collaboration
depends on sufficiently aligned internal representations of the task, the
equipment, the team members and their roles, and that these representations are
what let one party anticipate the other. Scheutz, DeLoach and
Adams~\cite{scheutz2017smm} observe that most of this literature quantifies the
benefit of such alignment while leaving open the data structures and processes
that would operationalize it, and propose a formal framework to fill that gap.
Transfiver can be read as one such operationalization at the level of a single
persistent computational state, with two differences worth stating. Alignment
here is not achieved by copying one party's model into the other, and the state
is not merely a shared view: it is the object both parties write to and the
object the model reads from.

\paragraph{States a person can see, and states a person can change.}
Concept bottleneck models let people edit predicted concepts and propagate the
change to outputs~\cite{koh2020concept}.  Concept Bottleneck Memory Models retain
past interventions and learn when to reapply them~\cite{steinmann2024cb2m}, making
them a particularly close comparison for persistent human correction.
Self-explaining networks build
faithfulness into model design~\cite{alvarez2018selfexplaining}.  Transfiver shares the commitment to intervention and to
faithfulness-by-design over post-hoc visualization. It differs by making the
faithful, intervenable object \emph{persistent and dynamically evolving} across
interactions, and by requiring all persistent interaction-specific information
that can affect future output to remain in that state.  Causal-abstraction work supplies a stricter language for
testing whether an interpreted high-level variable has the claimed causal role
under interchange interventions~\cite{geiger2023causal}; Transfiver's view/edit contract
should ultimately be evaluated at that standard rather than by decoding alone.

Reading an item's current standing by combining its own history with activation
from currently active neighbors is not new; it is the core of ACT-R's declarative
memory~\cite{anderson2004spreading}.
Transfiver's claim is architectural: the persistent state that supports such readings is
exposed to the user and is also the state on which future computation depends, so
changing that state changes future behavior.

\paragraph{Evaluation.}
LongMemEval identifies extraction, multi-session reasoning, temporal reasoning,
knowledge update, and abstention as central long-term-memory abilities
\cite{wu2024longmemeval}.  Transfiver adds architectural audits: whether the state that
appears correct is actually the state the system uses, and whether correction
takes effect without parameter changes.

\section{Transfiver Architecture: How Information Lives in Transfiver}
\label{sec:information-picture}

\subsection{One folder is not enough}

In Transfiver, an item is not limited to a single folder or context. The same item can belong to several projects without being copied. Items that look similar are not necessarily related, while very different items may be strongly connected by a real event or relationship. 
For example, a bank password and a server password are lexically similar, 
but should not be merged, whereas a particular server and its
random password may be lexically distant but joined by a specific event. Transfiver therefore represents both what an item means and how it is connected to other items.

\subsection{A full conceptual state}

A mature Transfiver realization may use

\begin{equation}
  S_t=(\{s_{t,i}\}_{i=1}^{N_t},R_t,\Gamma_t),
  \qquad
  s_{t,i}=(\kappa_i,v_{t,i},o_{t,i},\tau_{t,i},
         w_{t,i},u_{t,i},p_{t,i},\sigma_{t,i}),
  \label{eq:fullstate}
\end{equation}

where $R_t$ contains current relations, $\Gamma_t$ records relation or event history retained in the state, and the fields of $s_{t,i}$ denote identity, content, occupancy, event time, standing, use evidence, provenance, and lifecycle, respectively. Identity $\kappa_i$ is stable, while the remaining fields may change as the state evolves.

Relations are sparse.  Write $z_t$ for the encoding of the event arriving at step $t$; it is the only new information a step receives.  A contextual message may take the form

\begin{equation}
  m_{t,i}=G_\theta\!\left(z_t,s_{t,i},
  \{(r_{ij},s_{t,j}):(i,j)\in R_t\}\right).
  \label{eq:message}
\end{equation}

The message influences the current read.  It is not written into a neighbor
unless the transition observes a new event about that neighbor.  This keeps
derived influence reversible and attributable.

\subsection{The canonical interaction boundary}
\label{sec:interaction-boundary}

The state contract applies at the boundary where interaction enters Transfiver.  A
turn may contain user-authored text, a model statement, a tool or world result,
or text quoted or carried into a later user turn.  The raw turn and its local
context are retained in the event history.  Transport is evidence about how
text arrived. A user may quote an assistant,
approve it, reject it, or issue a new instruction next to it.  Interpretation
and uncertainty therefore belong to the learned transition and its readable
provenance, rather than to a hard-coded ``user text is true'' flag.

For an implementation to count as Transfiver, the canonical state must contain every
interaction-specific record that can affect a later output, including event
order, source, modality, relations, lifecycle changes, and edit records. The same
serialized state must be sufficient to restore the next transition and the
pre-decode output; a cache, hidden conversation window, or model-specific
history outside the serialization would violate the one-state contract.

\section{How Transfiver Learns and Changes}
\label{sec:learning}

\subsection{Two ways a state changes: directed edit and stream update}
\label{sec:two-changes}

A living state changes in two ways, and only one of them is exercised by an edit
interface.

In a \emph{directed edit} a person selects a visible item and supplies a new
value.  The address is chosen by the person, and the system must make the
correction take effect, leave unrelated items alone, and allow exact restoration.
This is what the protocols of Section~\ref{sec:one-state} measure.

In a \emph{stream update} no address is supplied.  An event arrives that concerns
something the state already holds: a password is replaced, a decision is
superseded, a claim is retracted.  Nothing marks the event as an update.  The
system itself must decide that it touches something already present rather than
introducing something new.

The two fail differently.  A system with a correct edit interface and no
stream-update capability does not accumulate memory: repeated mentions of the
same thing either leave no trace or create fresh items, and the state drifts
toward the condition described in the introduction, where nothing indicates which
copy is current.  Transfiver requires both, and they should be evaluated separately.

A judgment need not be made irreversibly at write time. When the current event alone does not determine whether an encounter is a genuine update or an incidental repetition, the transition can preserve the encounter and let later evidence determine how it should be read. The item is touched and can become the anchor for subsequent relations. A design that simply suppresses a write because the event resembles something already stored risks discarding a real update.

\subsection{Offline learning and online state evolution}

Transfiver separates learning the dynamics from using the learned dynamics.  During
offline training, future task loss must reach the transition:

\begin{equation}
  S_{t+1}=T_\theta(S_t,z_t),\qquad
  \theta\leftarrow\theta-\eta\nabla_\theta
  \sum_{t=1}^{H}\mathcal L(y_t,y_t^\star).
  \label{eq:offline}
\end{equation}

where $H$ is the training horizon.
During deployment, $\theta$ is fixed and stream updates evolve $S$ through the forward transition:

\begin{equation}
  \theta\ \text{fixed},\qquad S_{t+1}=T_\theta(S_t,z_t).
  \label{eq:online}
\end{equation}

Directed edits are the other route by which $S$ changes: they modify $S$
directly at an address the user supplies, without changing $\theta$. Neither
route touches the parameters. Thus ``learning from one event'' means persistent state change without gradient
retraining.

The constraint here is not that Transfiver avoids rules or avoids numbers.  \emph{No fixed value may
occupy the place of a judgment that belongs to the user}.

Would changing this
number change how the system computes, or would it change whose opinion the
system expresses?  A state dimension, a loss weight, or a layer count belongs to the first kind.  How far a source is trusted, when an item stops counting as live, what
outranks what: these are the second.  A rule might, for example, fuse an item's own trace with its neighbors' in a fixed proportion---seventy percent borrowed and thirty percent its own. Transfiver forbids such a fixed semantic weighting,
because those proportions differ across users, tasks, and histories, and drift
for one user over time.

Deterministic routing is not automatically a hand rule.  The test is where the
distinction comes from.  Whether a passage was typed by the user, emitted by a
model, or pasted from a document is a delivery address that arrives with the
input. A dictionary of marker words that
tries to decide from the text itself that something matters is a different
object: it recognizes surface form, and its coverage saturates.  Each extension
of such a dictionary buys recall with false positives, because the surface forms
of significance are open-ended in a way that channel of origin is not.

A new failure is not a license for a
new semantic flag; the first move is to ask whether an existing distinction,
read expressively enough, already yields the behavior, and to delete the
proposed concept when it does.

If Transfiver learns user-dependent judgments such as importance or source priority,
the learning signal must be specified.  One possible signal is subsequent user
behavior: whether an item is retrieved again, used in later work, or explicitly
marked as important.  The absence of such behavior is not treated as negative
evidence, since an unused item may still be valid or important.  Because a
single action may be noisy, repeated use provides stronger evidence than one
isolated interaction.

\subsection{Derive before adding}

The same constraint applies to architecture design.  A new failure should not
automatically create a new semantic flag, state field, or task-specific module.
The first question is whether the existing distinctions---content, context,
source, time, use, importance, and relation---produce the required behavior when
an expressive learned read combines them.  ``Dead,'' for example, need not be a
stored binary label if the relevant behavior follows from recency, continued
use, blocking events, importance, and surviving relations.  A false completion
need not be a special state kind if an unsupported self-report and a relevant
external receipt receive different standing.

Representing scope may require a learned pairwise or multiplicative comparison between a claim and evidence. Such an operator belongs to Transfiver only if it implements the shared
context/relation distinction and works beyond one completion benchmark.  A new
primitive is justified only after the existing state, equipped with a read
capable of expressing its own principles, fails the corresponding hostile tests.

\section{What Makes a System Transfiver?}
\label{sec:one-state}

If two copies of Transfiver have exactly the same saved state, then the same future input should induce the same distribution over future behavior. If those distributions differ, some information must still be stored somewhere outside the state the user can see and edit. We call this requirement \emph{history sufficiency}.

\subsection{The requirement: history sufficiency}

Let $H_t=(x_1,\ldots,x_t)$ denote everything that has happened up to time $t$. The state $S_t$ is what Transfiver has kept from that history. The shared parameters $\theta$ may contain general abilities learned before use, but they may not quietly store facts that belong to one user's interaction history.

Formally, if two histories leave Transfiver in the same state, no future continuation should be able to tell those histories apart. Here, $\operatorname{Law}(F_\theta(H_t,c))$ denotes the distribution over future behavior induced by history $H_t$ under continuation $c$. Importantly, $F_\theta$ denotes the behavior of the deployed system and is not assumed in advance to factor only through $S_t$.

\begin{equation}
S(H_t)=S(H'_t)
\Longrightarrow
\operatorname{Law}\!\left(F_\theta(H_t,c)\right)
=
\operatorname{Law}\!\left(F_\theta(H'_t,c)\right)
\quad\text{for every admissible continuation }c.
\label{eq:history-sufficiency}
\end{equation}

If this condition fails, some history-dependent information remains outside the state Transfiver shows.

Different applications may translate their inputs into different forms before Transfiver reads them, and may translate Transfiver's output into different forms afterwards. We write these temporary input and output adapters for a domain $d$ as $A_d^{\mathrm{in}}$ and $A_d^{\mathrm{out}}$. They may translate information, but they may not keep their own memory between interactions:

\begin{equation}
  (S_{t+1}, L_t, C_t)=
  \Step_\theta(S_t,A_d^{\mathrm{in}}(x_t)),
  \qquad
  y_t=A_d^{\mathrm{out}}(L_t),
  \label{eq:governing}
\end{equation}

Here, $\Step_\theta$ denotes one complete Transfiver step. Its state
component is the transition $T_\theta$ defined above: if
$\Step_\theta(S,z)=(S',L,C)$, then $S'=T_\theta(S,z)$.

Here, $S_t$ is the persistent state, $L_t$ is the model's output before the final decoding step, and $C_t$ records which state entries contributed to that output. The human-readable view is

\begin{equation}
  V_t=\View(S_t,C_t).
  \label{eq:view}
\end{equation}

The adapters may differ by domain, but they may not become a second hidden memory or a separate history-dependent route to the answer.

\subsubsection{Audit: nothing persistent outside the state}

Save $(\theta,S_t)$, restore it in a fresh process, and give the same next input. Under matched randomness, the restored system should reproduce the same next state and pre-decode output. It should induce the same distribution over future behavior.

\begin{definition}[Nothing persistent outside $S$]
Every interaction-specific persistent variable that can affect the distribution of future outputs is a member of $S_t$ and is included in serialization. Step-local queries, activations, and fresh random draws may be temporary, but interaction-specific information may not persist through an unlogged cache.
\end{definition}

\subsection{Optional: tracing which state entries produced the answer}

The small test model uses one extra restriction so that an answer can be checked entry by entry. For each active address, the implementation logs a contribution
\begin{equation}
  \widehat L_{t,i}
  =Q_\theta(z_t,s_{t,i},m_{t,i}),
  \qquad
  \widehat L_t
  =\sum_{i\in\mathcal I(S_t)}\widehat L_{t,i},
  \label{eq:logit-sum}
\end{equation}
where $L_t$ denotes the actual pre-decode logits produced by the
implementation and $\widehat L_t$ denotes the logits reconstructed solely
from the logged state-entry contributions.

Here, $\mathcal I(S_t)$ is the set of active addresses together with the unsupported sentinel $\bot$ of Section~\ref{sec:atomic}, whose contribution is logged like any other, and $\widehat L_{t,i}$ is the logged contribution from entry $i$. This lets us check whether an answer came from the visible state rather than from an unlogged route.

We audit path completeness with
\[
\epsilon_{\mathrm{path}}(t)
=
\|L_t-\widehat L_t\|.
\]
A system whose decoder receives no trainable history-dependent input outside
the logged contributions should have $\epsilon_{\mathrm{path}}(t)$ equal to
zero up to numerical precision.  In the small model of
Section~\ref{sec:atomic} the pre-decode logits are the sum of the logged
per-slot contributions and the logged sentinel contribution, so
$\epsilon_{\mathrm{path}}$ is zero by construction.

\subsection{Audit: what is shown must be what is used}

Let $g_\theta(S,x)$ denote Transfiver's one-step answer under state $S$ and input $x$. A successful edit should change the answer supported by the edited item, leave unrelated answers alone, and return to the original behavior when the edit is undone. Let $\Edit(S,i,q)$ replace the visible payload at address $i$ with $q$. Then we measure

\begin{align}
\mathrm{target}(i,q)
  &=\Pr[g_\theta(\Edit(S,i,q),x)=q\mid x\in X_i],\\
\mathrm{locality}(i,q)
  &=\mathbb E_{x\in X_{\neg i}}
    d(g_\theta(\Edit(S,i,q),x),g_\theta(S,x)),\\
\mathrm{rollback}(i,q)
  &=d_S(\Undo(\Edit(S,i,q)),S),\\
\mathrm{path}(t)
&=\left\|L_t-\widehat L_t\right\|.
\end{align}

The first quantity asks whether the target changed as intended. The second asks whether unrelated outputs changed. The third checks whether undo restores the original state. The last checks whether the logged contributions add up to the actual pre-decode output.

\subsection{Symmetry, lifecycle, and rollback}

If slot numbers have no meaning, simply reordering the slots should not change the model's behavior. For a permutation $P$, the next-state distribution should permute in the same way while the output distribution remains unchanged:

\begin{equation}
T_\theta(PS,x)\overset{d}{=}P T_\theta(S,x),\qquad
g_\theta(PS,x)\overset{d}{=}g_\theta(S,x),
\end{equation}

where $\overset{d}{=}$ denotes equality in distribution. This matters for lifecycle operations as well: retiring or restoring an item should depend on its state and history, not on an arbitrary storage index.

\begin{proposition}[Rollback restoration]
Assume fixed $\theta$ and history sufficiency. If an edit and its undo restore the serialized state exactly,
\[
\Undo(\Edit(S,i,q))=S,
\]
then an identical future continuation induces the same distribution over
future behavior as it did from $S$. Under matched randomness, a deterministic transition conditioned on that randomness reproduces the same trajectory.
\end{proposition}

\section{A Small Test Model}
\label{sec:atomic}

The current implementation is smaller than
Equation~\ref{eq:fullstate}.  Its purpose is to test whether a learned state can be the
sole persistent computational path while supporting semantic intervention.  It
uses

\begin{equation}
S=(K,V,O,A),
\end{equation}

with slot keys $K$, values $V$, occupancy $O$, and trainable empty-slot anchors
$A$.  There is no persistent recurrent hidden state.  The current event is
encoded into a step-local vector $h_t$, and a shared key encoder gives query
$q_t$.

For occupied slot $i$, addressing is

\begin{align}
b_{t,i}&=\langle \bar q_t,\bar K_i\rangle+\log(O_i+\epsilon),\\
(\alpha_{t,1:N},\alpha_{t,\bot})
  &=\operatorname{softmax}([b_{t,1:N},b_{t,\bot}]/\tau_{\mathrm{soft}}),
\end{align}

where a bar denotes $L_2$ normalization, $\bot$ is a learned unsupported sentinel and $\tau_{\mathrm{soft}}>0$ is the softmax temperature. Each slot produces

\begin{equation}
L_{t,i}=\alpha_{t,i}
W_o\left(\tanh(W_vV_i)\odot\tanh(W_qh_t)\right),
\end{equation}

and the pre-decode logits are exactly $\sum_iL_{t,i}+L_{t,\bot}$, where the
sentinel term $L_{t,\bot}$ is logged alongside the slot terms and is the only
source of the unsupported coordinate.

The transition learns a write probability and a hard forward address with a
straight-through gradient estimator.  A hierarchical router first chooses
between touching an existing identity and creating a new identity, then chooses
among available anchors.

The semantic interface is deliberately bounded.  It decodes each occupied slot
into a public categorical schema $(\text{key},\text{value})$ and permits a user
to replace one visible item.  It owns no persistent cache.  The edited key and
value are written into the canonical $K, V$ tensors, after which ordinary answers
continue through the same readout.  The fixed schema limits what the public interface can carry and makes exact
round-trip evaluation possible.

\section{Experiments and Results}
\label{sec:experiments}

\subsection{Real interaction records}
\label{sec:real}

The synthetic tests use a state we control. We now run the same
contract on real interaction records, where the state is built from a public
benchmark of multi-session human--assistant conversations in which a user states
a value and later replaces it. We keep the $44$ items in which two annotated
evidence turns carry an old and a new value for the same subject, place them in
the state, and let a learned read select four entries for the model.

We run three open models to separate what belongs to the store from what
belongs to the reader. We compare two ways of acting on a superseded entry. \emph{Prompt deletion} omits
the entry when assembling this turn's context, the operation available to any
retrieval system. \emph{State retraction} calls \texttt{retract} on the entry, so
the read no longer selects it while the entry remains in the state. Both arms show
the model the same number of entries, and the removed slot is refilled from the
same ranking, so neither arm wins by showing one line more.

\begin{table}[t]
\centering
\small
\begin{tabular}{lcccccc}
\hline
 & \multicolumn{2}{c}{Mistral-7B} & \multicolumn{2}{c}{Qwen2.5-7B} & \multicolumn{2}{c}{Phi-3-mini} \\
 & del. & retr. & del. & retr. & del. & retr. \\
\hline
Intervention: current value        & $.659$ & $.659$ & $.705$ & $.705$ & $.659$ & $.659$ \\
Recovering the earlier value       & $0$ & $17$ & $0$ & $14$ & $0$ & $17$ \\
Old entry returns next turn        & $31$ & $0$ & $31$ & $0$ & $31$ & $0$ \\
Current value next turn            & $19$ & $29$ & $23$ & $31$ & $20$ & $29$ \\
Restore after retraction           & n/a & $31$ & n/a & $31$ & n/a & $31$ \\
Same view after save and reload    & n/a & $44$ & n/a & $44$ & n/a & $44$ \\
\hline
\end{tabular}
\caption{The same contract on real records, under prompt deletion (del.) and
state retraction (retr.). The first row is accuracy; the remaining rows are
counts out of $44$ items. Four entries are shown, so chance is $.25$. A control that withholds both evidence turns answers $0/44$ in
every model. The three rows that
describe the state rather than the answer---return, restore, reload---are
identical across models, as they should be: they are properties of the store, not
of the reader.}
\label{tab:real}
\end{table}

First, on the current value the two arms are identical by
construction: they render the same lines, so a one-shot accuracy gap would be an
artifact rather than a result. Second, the arms separate on what happens next. A
prompt deletion changes nothing in the state, so on the following turn the old
entry returns in $31$ of $44$ items and next-turn accuracy falls below the
intervention level in every reader ($.432$--$.523$; for Mistral, $.659$ to
$.432$); a retraction holds, the entry never returns, and next-turn accuracy
stays at the intervention level ($.659$--$.705$). Third,
only the state arm can answer what the value used to be, because the retracted
entry is still there to be read when the question asks for it.

The recovered-earlier-value rate of $14$--$17$ out of $44$ is far
from the $31/44$ items in which the old entry is visible, so making an entry
available is not the same as using it. And our serialization carries content,
standing and relations; event times are restored in relative order, while
provenance, importance and usage counts are not yet serialized, so the
save--restore result covers the part of the state the read consumes.

\section{Limitations, Predictions, and Open Problems}

\subsection{Falsifiable predictions and proposed tests}

Table~\ref{tab:predictions} summarizes design-level predictions of Transfiver and corresponding tests that could falsify them. Some predictions are partially exercised by the synthetic experiments in this paper, while others remain untested.

\begin{table}[h]
\centering
\small
\caption{Falsifiable predictions and proposed tests. Each row states a predicted property and a construction that would refute it.}
\label{tab:predictions}
\begin{tabular}{p{0.27\linewidth}p{0.30\linewidth}p{0.35\linewidth}}
\toprule
Prediction & What must hold & Refuting construction\\
\midrule
No hidden persistent path & Holding $S$ fixed, no additional persistent interaction-specific variable changes future behavior & Restore the same $S$ in a fresh process; identify any unlogged persistent cache or history-dependent route that changes the output distribution\\
A forward obligation is state & What is still expected is held, not implied & Schedule an obligation; withhold the event; require the absence to be reported\\
Intention is not accomplishment & Plan, claim and receipt occupy different standing & Plan without receipt; claim without receipt; receipt without claim\\
One cause does not close a question & An explanation carries the scope in which it was established & Supply a sufficient-looking cause covering part of the observations only\\
User-marked importance remains distinct & Marked importance stays distinguishable from frequency and recency & Mark an item; bury it under unrelated volume and time; query it\\
Judgment follows a declared source & Review and translation bind to the source, not to fluency & Remove the source; provide a conflicting source; supply an unsupported source\\
Scope travels with what was learned & A rule learned in one project does not fire in another & Teach in project A; query in project B; re-scope; re-query\\
An instruction need not be repeated & Held instruction survives a session boundary & Instruct once; serialize; restore; require the behavior without restatement\\
Omission is visible & An unmet obligation appears as an absence in the view & Hold a list; satisfy all but one; require the missing one to be reported\\
What was learned is inspectable & The learned detail, not only the output, can be corrected & Train where a surface cue and a real quality both predict; ablate the cue; require the judgment to move\\
Deposits are cheaper than gradients & A fact placed in state costs less than a fact pressed into weights & Same fact, same budget: state deposit versus gradient acquisition\\
Quiet or dead need not be final & Retirement is reversible & Retire, distract, revive, partial inheritance, exact rollback\\
\bottomrule
\end{tabular}
\end{table}

\subsection{Open problems}

Scaling the architecture to free-form language and comparison with full-scale long-term-memory systems remain the next steps. A fuller realization should also incorporate human-readable visualization, source tracking, information liveness, complex relations, and reversible lifecycle in one learned system.

The architecture may create new failure modes.  If errors identified by humans are not corrected by humans, problems such as incorrect identity assignment, forced merging of different meanings, splitting of a single meaning into multiple, false source maintenance, self-read history contamination, and the discarding of quiet but important information may occur.

Finally, an architecture that makes user-specific state persistent creates privacy, security, and governance obligations. Provenance, deletion, export, access control, and user-visible uncertainty are part of a deployable system even when they are not all part of the learning architecture. Personalization can also adapt to harmful user behavior, creating safety risks beyond privacy alone.

\section{Conclusion}

Transfiver begins with a picture of information that cannot be reduced to tokens in one folder or values ordered only by recency. Its state is intended to jointly represent information that may be quiet but still alive, retired but recoverable, shared across projects, interpreted differently by different people, or changed through subsequent use.

The architectural claim of Transfiver is that one evolving state must remember, support
reasoning, produce output, expose what it contains, and accept correction.  The
history-sufficiency, intervention, and restoration requirements make
that claim falsifiable.  The small test model establishes only that part of this
contract can be built and audited in a narrow learned system.

The dissociations the broader architecture is intended to preserve---context
without fragmentation, multiple belonging without copying, relation without
similarity shortcuts, liveness without a recency shortcut, qualified observation
without self-amplification, and correction without a hidden bypass---are design
goals rather than results reported here. Testing them together in one system is
the work described under open problems.

\appendix
\section{Synthetic tests}
The following tests use a state we construct and a model we train, so they
establish that parts of the contract are satisfiable rather than that the whole
contract holds in the wild. The real-record results in Section~\ref{sec:real} carry that claim.

\subsection{Contexts and overlap}
\label{sec:context}

We test whether the same content can stay separate when it is placed in two
different contexts.  The address here is formed in two halves, one from the key
and one from the context; nothing else about the model changes.  An episode writes four facts on context~0 and the same four
keys on context~1, giving two of those keys deliberately different values in the
two contexts; the events of the two contexts are interleaved in random order.
The query names a key \emph{and} a context, and the correct answer is the value
that key holds in the named context, not the value written most recently.  Two
conditions differ in how the context reaches the model.  In the easier
condition the context is a field on every event.  In the harder condition a
single header event announces the context and the following events carry no
context field at all, so the current context has to be carried forward in the
state.  Values are 16-way, so chance is $1/16=.0625$ and the pre-specified
threshold is chance $+\,3\text{SE}=.095$; each seed is evaluated on 512 queries
from episodes generated with an evaluation seed disjoint from training.

\paragraph{How each number is measured.}
\begin{description}[leftmargin=1.2em,itemsep=1pt,topsep=2pt]
  \item[Context-conditioned accuracy] Fraction of evaluation queries whose
  answer, the arg-max over the 16 value coordinates, equals the value the
  queried key holds in the named context.
  \item[Recency collapse] Fraction whose answer equals the value the same key
  holds in the \emph{other} context.  By construction the two differ, so this
  isolates one specific failure: answering from the other, in general more
  recent, binding rather than from the one asked for.
  \item[Flipping the queried context returns the other value] The state is left
  untouched and only the context coordinate of the query is switched; the
  fraction whose answer becomes the other context's value.  This is the causal
  form of the first row: the context in the query, not the key alone, selects
  the answer.
  \item[Same key placed on different items] The read attends over slots; this is
  the fraction of queries for which the arg-max slot for (key, context~0)
  differs from the arg-max slot for (key, context~1) -- that is, the two
  bindings occupy different entries instead of one entry that was overwritten.
  \item[Context withdrawal changes the prior answer] Every slot whose context axis is
  nearest to context~0 has its occupancy set to zero; contents are not touched.
  Fraction of queries whose context~0 answer then differs from what it was
  before.  It measures that the answer changes, not that the model reports the
  entry as unsupported; the stricter form is reported separately below.
  \item[Restoring the context returns it exactly] The saved occupancy vector is
  written back into the withdrawn state; the fraction whose answer equals the
  pre-withdrawal answer, together with a check that the restored state is
  tensor-identical to the original.
  \item[Overlap] The overlap between two contexts is computed at read time as a
  max-match on the content
  half of the slot keys: for every item placed on $c$, the cosine similarity to
  the most similar item placed on $c'$, averaged over the items of $c$ (self
  pairs excluded when $c=c'$).  Nothing is stored in either context; the value
  exists only while the view is drawn.
  \item[Items created per episode] Mean number of occupied slots at the end of
  an episode; eight facts written on two contexts should occupy eight entries if
  the two contexts do not merge, and nine in the second condition, where the
  header event opens an entry of its own.
\end{description}

\begin{table}[H]
\centering
\caption{Context test, five seeds, 512 evaluated queries per seed.  The same key
appears in two contexts with different values, so the value the key holds in the
other context is always a distinct wrong answer.  Chance is $1/16=.0625$ and the
pre-specified threshold is chance $+\,3\text{SE}=.095$.  In the second block the
context is announced once by a header event and must be carried forward in the
state.}
\label{tab:context-gate}
\small
\begin{tabular}{lccccc}
\toprule
Metric & Seed 0 & Seed 1 & Seed 2 & Seed 3 & Seed 4\\
\midrule
\multicolumn{6}{l}{\emph{Context given on every event}}\\
Context-conditioned accuracy & 1.000 & 1.000 & .486 & 1.000 & 1.000\\
Recency collapse (lower is better) & .000 & .000 & .436 & .000 & .000\\
Flipping the queried context returns the other value & 1.000 & 1.000 & .486 & 1.000 & 1.000\\
Same key placed on different items & 1.000 & 1.000 & .135 & 1.000 & 1.000\\
Withdrawing the context changes its prior answer & 1.000 & 1.000 & .547 & 1.000 & 1.000\\
Withdrawn context answered as unsupported & .000 & .000 & 1.000 & .000 & .000\\
Restoring the context returns it exactly & 1.000 & 1.000 & 1.000 & 1.000 & 1.000\\
Overlap, different contexts & .999 & 1.000 & .190 & 1.000 & .999\\
Overlap, within one context & .149 & .135 & .070 & .152 & .138\\
Items created per episode & 8.0 & 8.0 & 3.9 & 8.0 & 8.0\\
\midrule
\multicolumn{6}{l}{\emph{Context announced once, then carried in the state}}\\
Context-conditioned accuracy & 1.000 & 1.000 & 1.000 & 1.000 & 1.000\\
Recency collapse (lower is better) & .000 & .000 & .000 & .000 & .000\\
Flipping the queried context returns the other value & 1.000 & 1.000 & 1.000 & 1.000 & 1.000\\
Same key placed on different items & 1.000 & 1.000 & 1.000 & 1.000 & 1.000\\
Withdrawing the context changes its prior answer & 1.000 & 1.000 & 1.000 & 1.000 & 1.000\\
Withdrawn context answered as unsupported & .000 & .000 & .000 & .000 & .000\\
Restoring the context returns it exactly & 1.000 & 1.000 & 1.000 & 1.000 & 1.000\\
Overlap, different contexts & .891 & .904 & .900 & .885 & .886\\
Overlap, within one context & .165 & .177 & .154 & .148 & .169\\
Items created per episode & 9.0 & 9.0 & 9.0 & 9.0 & 9.0\\
\bottomrule
\end{tabular}
\end{table}

Four of the five seeds in the first condition and all five in the second reach
a solution in which the two contexts stay separate: the same key occupies two entries, the answer
follows the context named in the query rather than the most recent write, and
switching only the context coordinate of the query returns the other context's
value.  Withdrawing a context by clearing the occupancy of the items placed on
it changes the answer previously supported by that context, and writing the
saved occupancy back reproduces the previous answer from a state that is
tensor-identical to the original; the items themselves are never edited.  After withdrawal the model answers from the remaining context and never routes
to the sentinel $\bot$ of Section~\ref{sec:atomic}, so the ``unsupported'' row
of Table~\ref{tab:context-gate} is $.000$ in every seed that separates the
contexts.  Abstention on withdrawn evidence, which LongMemEval lists as a core
ability~\cite{wu2024longmemeval}, is therefore not established by this
experiment.

One seed of five does not reach this solution in the first block.  It opens
$3.9$ entries per episode instead of eight, which is consistent with the two
contexts being merged into shared entries, and the other quantities move with
that failure: context-conditioned accuracy falls to $.486$, the other context's
value is returned in $.436$ of queries, and cross-context overlap falls from
near one to $.190$.  The same seed succeeds in the second block, where the
context has to be carried in the state.  Five seeds do not establish how often the
solution is reached: as a sign test, four successes against one failure gives
$p=.375$, and five against none gives $p=.0625$.

\subsection{Reading four rules out of one state}
\label{sec:readout}

A necessary condition for a state to expose multiple temporal distinctions is
that those distinctions remain recoverable from the same frozen state.  This
section tests that condition on an earlier implementation of a persistent state
built from the same commitments---items held in slots, joined by relations that
carry the time and the count of the events that formed them---using a single
held-out evaluation fixed in advance.  It is not the model of
Section~\ref{sec:atomic}, whose slots hold a key, a value and an occupancy and
no relation history, so this question cannot be put to that model.

\paragraph{The task.}  An episode is 512 tokens long.  Fifty-eight keys are each
written four times, interleaved with unrelated filler, and the episode is
divided into eight numbered sections that act as coarse timestamps.  Each
episode ends with eight queries.  A query names a key and one of four rules:

\begin{enumerate}[leftmargin=1.4em,itemsep=1pt,topsep=2pt]
  \item \emph{most recent} --- the last value written for that key;
  \item \emph{earliest} --- the first value written for that key;
  \item \emph{as of a named section} --- the value that was in force just before
  a named section of the episode;
  \item \emph{the $n$th value} --- the $n$th value ever written for that key.
\end{enumerate}

The last two rules carry a number with them, and in this evaluation that number
is always the second or the third.  All four rules appear in the same episode,
two queries each, so the same key can be asked about under different rules and
the four rules disagree with one another; the text of the episode alone does not
determine the answer.  Which values a key may take is fixed by a hash into three
disjoint pools, one used while training, one while selecting, and one reserved
for the held-out test, so the values that appear as answers in the final test
never appeared as answers earlier.

\paragraph{The readout.}  The trained model is frozen.  For a query about a key,
the \emph{candidates} are the $205$ values that key is allowed to take in this
pool; four of them were actually written.  From the frozen state we take five
numbers per candidate, and nothing else:

\begin{enumerate}[leftmargin=1.4em,itemsep=1pt,topsep=2pt]
  \item whether the state currently holds a relation from the queried key to
  that candidate;
  \item how long before the query that relation was formed, as a fraction of the
  episode;
  \item where that relation falls in time among the key's own relations, as a
  rank scaled to $[0,1]$;
  \item the numbered section in which it was formed, minus the section the query
  named;
  \item its rank in time, minus the number the query named.
\end{enumerate}

The last two are zero for the two rules that name no number, which is how the
named number reaches the readout at all.  Raw keys, raw values, and absolute
positions in the episode are excluded by construction and recorded as excluded.
These five numbers, together with a four-way indicator naming which rule is in
force, go to one small network of 385 parameters---capped at 400 in advance---
which scores each candidate separately; the answer is the highest-scoring
candidate.  The same network serves all four rules.  The four-way indicator
identifies which rule to apply, while the final two features carry the numeric
reference for the two rules that require one.  Each rule is a short arithmetic
function of these five numbers---extremes of the time rank for the first two,
the signed section difference for the third, the signed rank difference for the
fourth---so what the readout has to find in the state is those five quantities,
kept separable, rather than a rule.  The readout is trained on one
set of episodes, selected on a second, and the test partition was fixed before
evaluation and opened once, with retries and regeneration forbidden.

\paragraph{Result.}  Table~\ref{tab:readout} reports the held-out test over five
independently trained models, $1{,}280$ queries per rule and $5{,}120$ in total.
Guessing uniformly among a key's $205$ permitted values scores $.005$, but the
first feature already separates the four values written for that key from the
other $201$, so the baseline the table should be read against is one of those
four, $.250$.

\begin{table}[H]
\centering
\caption{Held-out readout test, five trained models, $1{,}280$ queries per rule.
The upper block is accuracy under each rule; the lower blocks are the controls
that were required to pass for the result to count.  In the two substitution
controls the informative number is not that accuracy on the originally named
rule falls to zero, but that the readout then answers the substituted rule
instead, which separates ``it stopped working'' from ``it followed the
instruction it was given''.}
\label{tab:readout}
\small
\begin{tabular}{lcccc}
\toprule
& Most recent & Earliest & As of a section & The $n$th value\\
\midrule
Accuracy & 1.000 & 1.000 & .984 & .996\\
\midrule
\multicolumn{5}{l}{\emph{The named rule is replaced by a different rule}}\\
Accuracy on the rule originally named & .000 & .000 & .004 & .000\\
Answers the substituted rule instead & 1.000 & 1.000 & .996 & .984\\
\midrule
\multicolumn{5}{l}{\emph{The named number is replaced} (only the two rules that carry one)}\\
Accuracy on the number originally named & --- & --- & .000 & .008\\
Answers the substituted number instead & --- & --- & .980 & .992\\
\midrule
\multicolumn{5}{l}{\emph{Invariance and leakage}, over all $5{,}120$ queries}\\
\multicolumn{5}{l}{Every key renamed: predictions identical, largest change in score $0.0$}\\
\multicolumn{5}{l}{Items moved within the episode: predictions identical, largest change in score $0.0$}\\
\bottomrule
\end{tabular}
\end{table}

These results show that the temporal distinctions required by the four rules
remain recoverable from the frozen state.  They do not show that the model's own
answer path uses the same readout.  The task is also synthetic: keys, values,
and sections are symbols, and the two rules that name a number were tested at
two of the eight sections rather than across all of them.

\bibliographystyle{plain}
\bibliography{references}

\end{document}